\documentclass[10pt,twocolumn,letterpaper]{article}

\usepackage{iccv}
\usepackage{times}
\usepackage{epsfig}
\usepackage{graphicx}
\usepackage{amsmath}
\usepackage{amssymb}
\usepackage{cite}
\usepackage{multirow}
\usepackage{color}
\usepackage{bm}
\usepackage{enumitem}
\usepackage{wrapfig}
\usepackage{float}
\usepackage[noend]{algpseudocode}
\usepackage{algorithmicx}
\usepackage{algorithm}

\usepackage[breaklinks=true,bookmarks=false]{hyperref}

\iccvfinalcopy 

\def\iccvPaperID{10990} 

\ificcvfinal\fi

\begin{document}

\title{CAM-loss: Towards Learning Spatially Discriminative Feature Representations}

\author{Chaofei Wang$^*$, Jiayu Xiao\thanks{Equal contribution.} \,, Yizeng Han, Qisen Yang, Shiji Song, Gao Huang\thanks{Corresponding author.}	\\
Department of Automation, Tsinghua University\\
}

\maketitle
\ificcvfinal\thispagestyle{empty}\fi

\begin{abstract}
The backbone of traditional CNN classifier is generally considered as a feature extractor, followed by a linear layer which performs the classification. We propose a novel loss function, termed as CAM-loss, to constrain the embedded feature maps with the class activation maps (CAMs) which indicate the spatially discriminative regions of an image for particular categories. CAM-loss drives the backbone to express the features of target category and suppress the features of non-target categories or background, so as to obtain more discriminative feature representations. It can be simply applied in any CNN architecture with neglectable additional parameters and calculations. Experimental results show that CAM-loss is applicable to a variety of network structures and can be combined with mainstream regularization methods to improve the performance of image classification. The strong generalization ability of CAM-loss is validated in the transfer learning and few shot learning tasks. Based on CAM-loss, we also propose a novel CAAM-CAM matching knowledge distillation method. This method directly uses the CAM generated by the teacher network to supervise the CAAM generated by the student network, which effectively improves the accuracy and convergence rate of the student network.
\end{abstract}

\section{Introduction}

\begin{figure}[t]
	\begin{center}
		\includegraphics[width=0.99\linewidth]{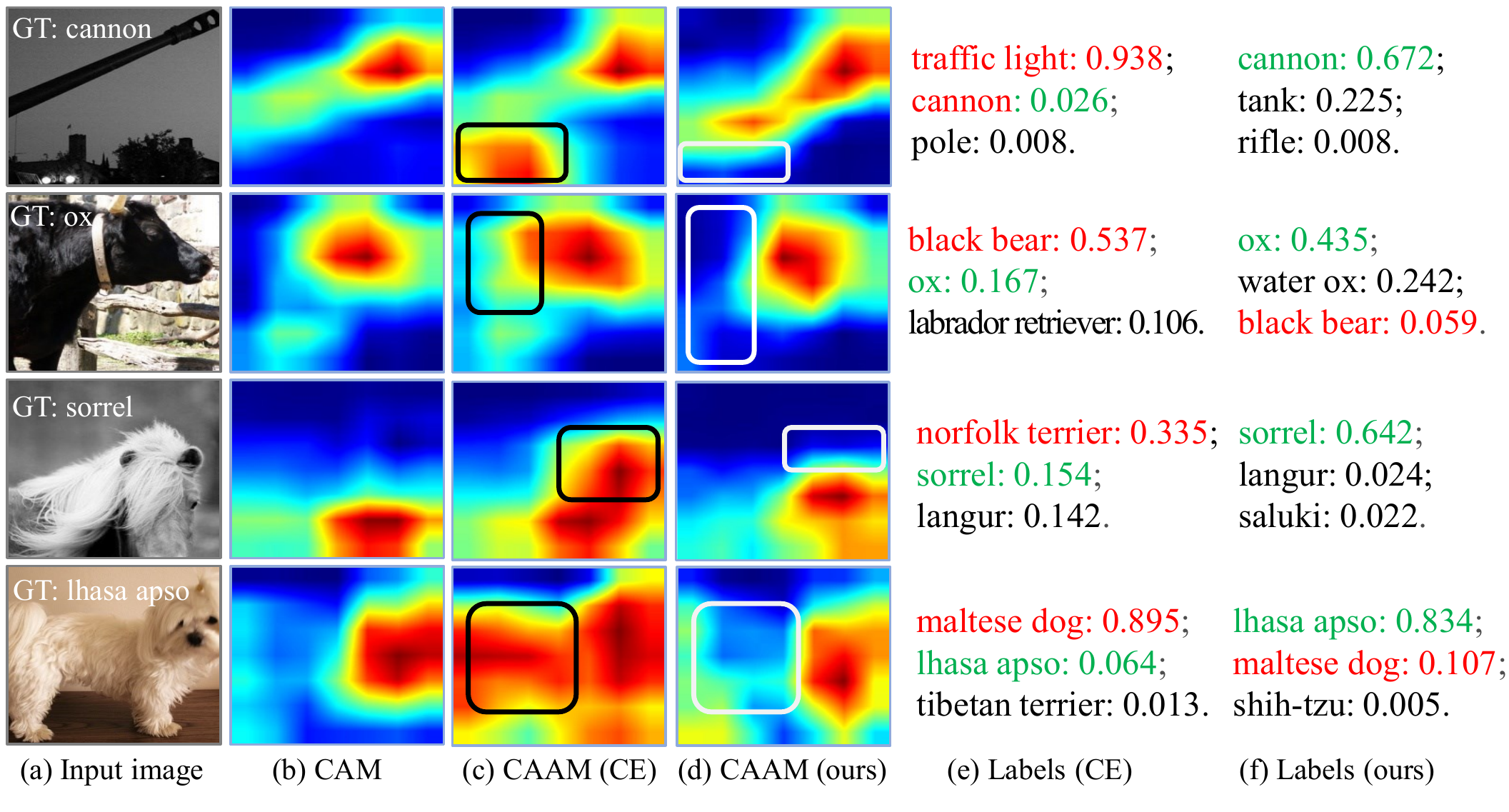}
	\end{center}
    \vspace{-1ex}
	\caption{Some examples to illustrate our motivation. A ResNet-50 model trained on ImageNet is adopted. ``GT'' represents the ground truth label. ``CE'' represents cross entropy loss. ``ours'' represents CAM-loss. Black bounding boxes show main differences between (b) and (c), while white ones show main differences between (c) and (d).}
	\vspace{-2ex}
	
	\setlength{\belowcaptionskip}{-0.3cm}
	\setlength{\abovecaptionskip}{-0.2cm}
	\label{fig:intuition}
\end{figure}
In the past few years, convolutional neural networks (CNNs) have achieved excellent performance in many visual classification tasks. To handle the increasingly complex data, CNNs have continuously been improved with deeper structures (AlexNet\cite{krizhevsky2012imagenet}, VGGNet\cite{simonyan2014very}, ResNet\cite{he2016deep}, ResNext\cite{xie2017aggregated}, DenseNet\cite{huang2017densely}). However, deep networks are prone to overfitting while they get stronger learning ability. Many researchers have proposed effective regularization solutions, such as Dropout\cite{srivastava2014dropout}, Weight Decay\cite{goodfellow2016deep}, Stochastic Depth\cite{huang2016deep}, Mixup\cite{zhang2017mixup}, Shakedrop\cite{yamada2018shakedrop}, Cutmix\cite{yun2019cutmix}. An alternative solution is to design different loss functions to obtain more distinguishing feature representations, which increase intra-class compactness and inter-class separability. Inspired by such idea, contrastive loss\cite{hadsell2006dimensionality}, triplet loss\cite{schroff2015facenet}, center loss\cite{wen2016discriminative} are proposed to bring in additional constraints on the basis of cross entropy loss. Unfortunately, they usually dramatically increase the computational cost. L-Softmax\cite{liu2016large} and SM-Softmax\cite{liang2017soft} are proposed to modify the original softmax function mathematically, leading to potentially larger angular separability between feature vectors. Implicit semantic data augmentation (ISDA) loss \cite{wang2021regularizing} is proposed to optimize the upper bound of expected cross-entropy loss. However, when adopting the above loss functions, an input image is represented by a one-dimensional feature vector, which collapses the spatial information.

In this paper, we propose to construct a novel loss function by leveraging the class activation maps (CAMs\cite{zhou2016learning}) with rich spatial information. CAM indicates the spatial discriminative regions to identify a particular category. It is easily obtained by computing a weighted sum of the feature maps of the last convolutional layer. In fact, we can also obtain a class-agnostic activation map (CAAM\cite{baek2020psynet}) by computing the sum of the feature maps directly, which indicates the spatial distribution of the embedded features. To describe our motivation visually, in \autoref{fig:intuition} we show some validation images misclassified by a pre-trained ResNet-50\cite{he2016deep} model with cross entropy loss on ImageNet\cite{russakovsky2015imagenet}. The CAMs of target categories, CAAMs and output labels are shown in columns (b), (c), and (e), respectively. A visual conclusion is that CAAMs generally show larger activated areas and richer features than CAMs of target categories. Unfortunately, the redundant feature representations (black bounding box areas in column (c)) result in the confidence scores of non-target categories (red labels in column (e)) exceeding those of target categories (green labels in column (e)), which lead to misclassification. For example, the expression of body makes the model misidentify an ox as a black bear, and the expression of ears and mane makes the model misidentify a horse as a dog. Intuitively, if we constrain CAAMs closer to CAMs of target categories, features of the target categories will be expressed well and those of non-target categories will be suppressed simultaneously. This effectively enforces intra-class compactness and inter-class separability.


Based on the above inspiration, we construct a new loss function, termed as CAM-loss, by minimizing the distance between the CAAM and the CAM of target category for each training image. CAM-loss drives the backbone to learn more discriminative feature representations from a spatial perspective. We train another ResNet-50 model with CAM-loss, and show CAAMs and output labels of the same images in \autoref{fig:intuition} (d) and (f). It shows that CAAMs produced by CAM-loss are usually cleaner than those produced by cross entropy loss (comparing column (c) with column (d)). Some features of non-target categories are suppressed (white bounding box areas in column (d)), which greatly improves the accuracy of labels (comparing column (e) with column (f)). In fact, extensive experiments show that CAM-loss effectively improves the performances of various classification models.

As an independent loss module, CAM-loss can also be combined with the mainstream regularization methods to improve their performances. Furthermore, we verify the strong generalization ability of CAM-loss in transfer learning and few shot learning tasks. CAM-loss particularly boosts the baseline method by $7.04\%$ (1-shot) and $4.75\%$ (5-shot) on CUB\cite{wah2011caltech}, $2.78\%$ (1-shot) and $1.68\%$ (5-shot) on Mini-ImageNet\cite{vinyals2016matching} in the setting of few shot learning. This is attributed to the key role of CAM-loss in reducing the negative effect of image background.

In the traditional teacher-student knowledge distillation framework, the existing methods all use a certain type of teacher knowledge to supervise the same type of student knowledge, such as soft target\cite{hinton2015distilling}, hints\cite{romero2014fitnets}, attention map\cite{zagoruyko2016paying}, relationship between samples\cite{peng2019correlation} or layers\cite{yim2017gift}. Inspired by CAM-loss, we propose a different idea to match different types of knowledge between teacher and student, that is, to directly supervise the CAAMs generated by the student network with the CAMs generated by the teacher network. We call it CAAM-CAM matching (CCM) knowledge distillation. The experimental results show that CCM can effectively improve the accuracy and convergence rate of the student network.

The main contributions of our work are:
\begin{itemize}[itemsep= 0 pt,topsep = 0 pt, parsep = 0 pt]
	\item We propose a novel loss function CAM-loss from the perspective of spatial information. It can effectively improve the classification performance of various CNN models with neglectable additional parameters and calculations, and easily be combined with the mainstream regularization methods to achieve the state-of-the-art on CIFAR-100 and ImageNet.
	\item CAM-loss shows strong generalization capability in transfer learning and few shot learning tasks. In particular, CAM-loss significantly improves the performance of few shot image classification.
	\item We propose a novel knowledge distillation method named CAAM-CAM matching, which matches different types of knowledge between teacher (CAMs) and student (CAAMs), and improves the accuracy and convergence rate of the student network simultaneously. 
\end{itemize}

\section{Related work}
\label{gen_inst}

\subsection{Class Activation Map}

Generating class activation maps (CAMs\cite{zhou2016learning}) with CNN classification models plays an important role in computer vision. Grad-CAM\cite{selvaraju2016grad} and Grad-CAM++\cite{chattopadhay2018grad} generalize CAM\cite{zhou2016learning}, so that CAMs can be obtained in any CNN-based classification models. The CAM technique derived from the classification network has been widely used for other weakly supervised visual tasks, such as localization\cite{yang2020combinational,baek2020psynet}, detection\cite{wan2018min,zhang2018zigzag,wei2018ts2c}, segmentation\cite{wei2018revisiting,li2018tell,ahn2019weakly}. 

In image classification tasks, CAM is usually used as a visualization technique, but few researchers treat it as something that can be fed back into training \cite{fukui2019attention,guo2019visual,sun2020fixing}. \cite{fukui2019attention} introduced a complicated multi-branch structure consisting of an attention mechanism, an attention branch (based on CAM), and a perception branch. \cite{guo2019visual} used dual image input and two-branch structure to do the attention consistency (based on CAM). These two methods rely on complex network structures and result in additional computation cost. In contrast, CAM-loss directly introduces the distance constraint between CAAM and CAM, both of which are generated in the normal training process of the classification model. Our method is very clean, direct, low-cost but effective. \cite{sun2020fixing} proposed to suppress the features of negative categories by minimizing the CAMs of top-k negative categories or constraining them with a uniform spatial distribution. Compared with \cite{sun2020fixing}, CAM-loss suppresses more extensive non-target regions such as the background. Furthermore, because the CAM of the target category and those of the non-target categories may overlap in some regions, CAM-loss can better avoid the risk of simultaneously suppressing the features of the target category.

\subsection{Loss function}
The cross entropy loss is widely used in CNNs due to its simplicity and probabilistic interpretation. Despite its popularity, it does not explicitly encourage the intra-class compactness and inter-class separability. One solution route is to add an additional loss term to assist the cross entropy loss. The contrastive loss\cite{hadsell2006dimensionality} was proposed to simultaneously minimize the distances between positive image pairs and enlarge the distances between negative image pairs. Similarly, the triplet loss\cite{schroff2015facenet} was proposed to apply a similar strategy to image triplets rather than image pairs. The center loss\cite{wen2016discriminative} was proposed to minimize the euclidean distance between the feature vector and the corresponding class centroid. A major drawback of these losses is the expensive calculation on image pairs or triplets explosion, class centroids update. Another solution route is to modify the softmax cross entropy loss. L-Softmax\cite{liu2016large}, SM-Softmax\cite{liang2017soft} and AM-Softmax\cite{wang2018additive} were proposed to introduce some margin parameters into the softmax function. ISDA\cite{wang2021regularizing} was proposed to optimize the upper bound of expected cross-entropy loss. However, in these methods, images are represented by one-dimensional feature vectors, which do not include any spatial information.

Different from the previous methods, CAM-loss utilizes the spatial information of self-generated CAMs to constrain the feature maps, which drives the backbone of CNN to learn more spatially discriminative feature representations. It has notable visual interpretability, and requires little additional computation. These advantages make it suitable for a wide range of application scenarios.

\subsection{Knowledge distillation}
A vanilla knowledge distillation (KD \cite{hinton2015distilling}) proposed to transfer some knowledge of a strong capacity teacher model to a compact student model by minimizing the Kullback-Leibler divergence between the soft targets of two models. Since then, there have been works exploring variants of knowledge distillation. Fitnets \cite{romero2014fitnets} proposed to transfer the knowledge using not only final outputs but also intermediate ones. AT\cite{zagoruyko2016paying} proposed an attention-based method to match the activation-based and gradient-based spatial attention maps. FSP\cite{yim2017gift} proposed to compute the Gram matrix of features across layers for knowledge transfer. CCKD\cite{peng2019correlation} proposed to transfer the correlation between instances. Existing methods all use a certain type of teacher knowledge to supervise the same type of student knowledge. Different from them, we first propose a new idea to match different types of knowledge between teacher and student, which can effectively improve the accuracy and convergence rate of the student network simultaneously.

\section{Method}
\label{headings}

In this section, we first formally define and describe the proposed CAM-loss, then analyze the choice of the hyper parameters, finally introduce and explain CAAM-CAM matching knowledge distillation.

\subsection{Definition of CAM-loss}
\begin{figure*}[t]
	\begin{center}
		\includegraphics[width=0.99\linewidth]{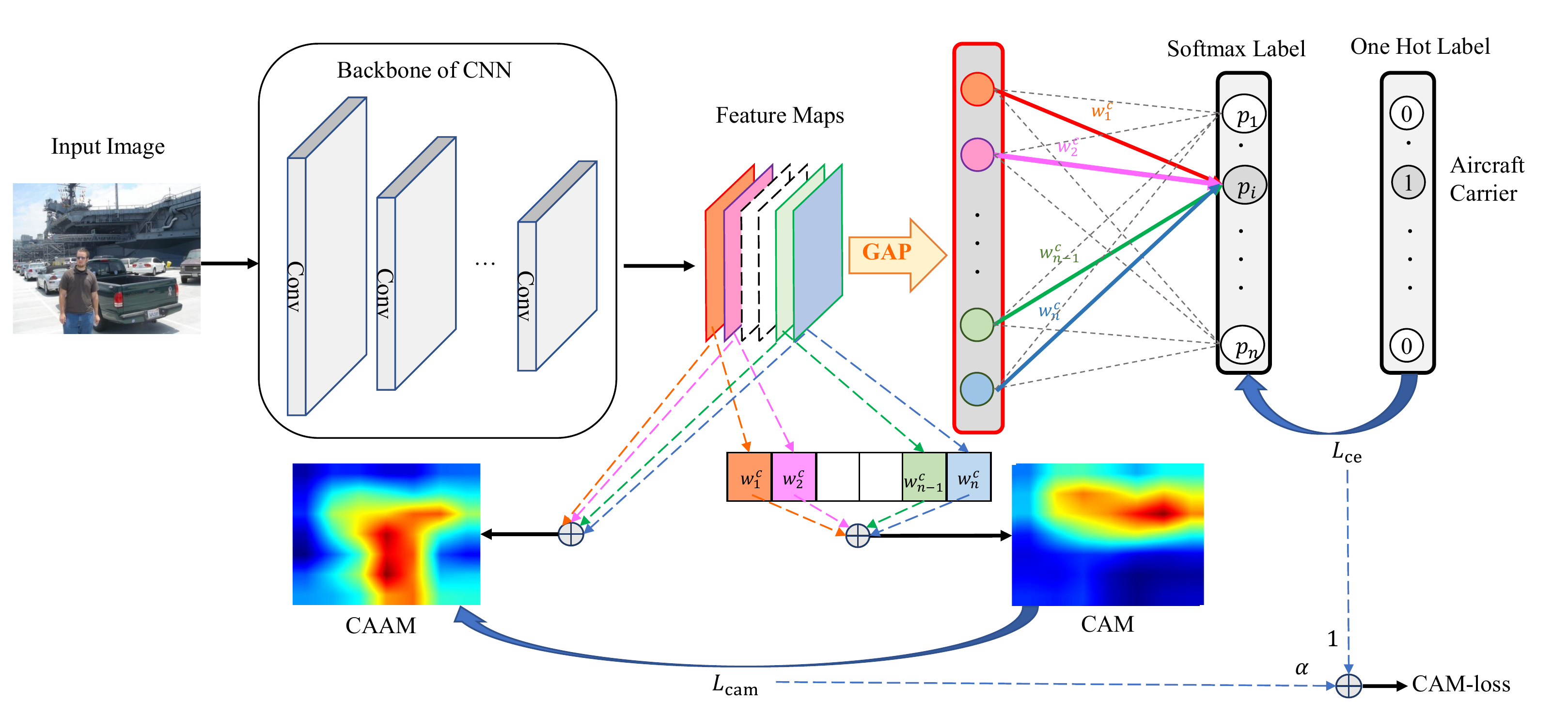}
	\end{center}
	\caption{How to get the CAM, CAAM and CAM-loss in a CNN framework. CAM is a weighted sum of the feature maps of the last convolutional layer. CAAM is the sum of the feature maps directly. CAM-loss is the combination of $L_{\text{cam}}$ and $L_{\text{ce}}$}
	\vspace{-2ex}
	\label{fig:structure}
\end{figure*}

Based on the procedure of generating CAMs in\cite{zhou2016learning}, we present how to get the CAM, CAAM, and CAM-loss in a CNN-based architecture as shown in \autoref{fig:structure}. Note that we can also use the generalized methods Grad-CAM\cite{selvaraju2016grad} or Grad-CAM++\cite{chattopadhay2018grad} to replace the method of CAM\cite{zhou2016learning}. The only difference is that, due to the calculation of gradients, Grad-CAM\cite{selvaraju2016grad} or Grad-CAM++\cite{chattopadhay2018grad} will increase the computational cost. In the paper, the method of CAM\cite{zhou2016learning} is chosen for the convenience of description and reduction of experiment cost. The formal description is shown below.

For a given image, the last convolutional layer outputs some units of feature map. Let  $f_{k}(x, y)$ represents the activation of unit $k$ at spatial location $(x, y)$. Then, for unit $k$ of height $H$ and width $W$, the result of performing global average pooling, $F_{k}= \frac{1}{H\times W}\sum _{x, y}f_{k}(x, y)$. Thus, for a given class $i$, the input to the softmax, $z_{i}= \sum _{k}w_{k}^{i}F_{k}$, where $w_{k}^{i}$ is the weight corresponding to class $i$ for unit $k$. Essentially, $w_{k}^{i}$ indicates the importance of $F_{k}$ for class $i$. Finally the output of the softmax for class $i$, $p_{i}$ is given by $\frac{e^{(z_{i})}}{\sum _{j}e^{(z_{j})}}$. By plugging $F_{k}= \frac{1}{H\times W}\sum _{x, y}f_{k}(x, y)$ into the class score $z_{i}$, we obtain

\begin{equation}
\begin{aligned}
z_{i} &= \frac{1}{H\times W}\sum _{k}w_{k}^{i} \sum _{x, y}f_{k}(x, y) \\ &= \frac{1}{H\times W}\sum _{x, y} \sum _{k} w_{k}^{i}f_{k}(x, y).
\end{aligned}
\end{equation}
We define $\text{CAM}_{i}$ as the class activation map for class $i$, where each spatial element is given by

\begin{equation}
\text{CAM}_{i}(x, y)= \sum _{k} w_{k}^{i}f_{k}(x, y).
\end{equation}
Thus, $z_{i} = \frac{1}{H\times W}\sum _{x, y}\text{CAM}_{i}(x, y)$, where $\text{CAM}_{i}(x, y)$ directly indicates the importance of the activation at spatial location $(x, y)$ leading to the image belonging to class $i$.

Furthermore, we define CAAM as the class-agnostic activation map for the input image. Each spatial element of CAAM is given by 

\begin{equation}
\text{CAAM}(x, y)= \sum _{k} f_{k}(x, y).
\end{equation}
To drive CAAM close to $\text{CAM}_{i}$, we define $L_{\text{cam}}$ to measure the distance between CAAM and $\text{CAM}_{i}$. After the same min-max normalization of CAAM and $\text{CAM}_{i}$, we get $\text{CAAM}^{'}$ and $\text{CAM}_{i}^{'}$, and then use any pixel space distance to measure $L_{\text{cam}}$. In this paper, we simply choose $l_{1}$ distance ($l_{2}$ could be used as well). So, the formal expression of $L_{\text{cam}}$ is given by
\begin{small}
	\begin{equation}
	L_{\text{cam}}\!=\!\frac{1}{H\times W}\sum _{x,y}\left \| \text{CAAM}^{'}(x,y) \!-\! \text{CAM}_{i}^{'}(x,y) \right \|_{l_{1}}.
	\end{equation}
\end{small}
Of course, the cross entropy (CE) loss $L_{\text{ce}}$ is still necessary and defined as follow:
\begin{equation}
L_{\text{ce}}=-\text{log}\frac{e^{(z_{i})}}{\sum _{j}e^{(z_{j})}}.
\label{equ:ce}
\end{equation}
When updating parameters of backbone, the two loss terms should be well combined as follow
\begin{equation}
\text{CAM-loss} =\alpha L_{\text{cam}} + L_{\text{ce}},
\label{equ:camloss}
\end{equation}
where $\alpha$ represents the combine ratio.

The training process is summarized in algorithm \autoref{alg:cam}. Note that $L_{\text{ce}}$ is used to update $W$ while CAM-loss is used to update $\theta$. The purpose is to eliminate the correlation between $W$ and $L_{\text{cam}}$, which may cause $W$ to approach a all one vector, resulting in an illusory decline of CAM-loss.
\begin{algorithm}[h]
	
	\caption{Training process with CAM-loss} 
	\hspace*{0.02in} {\bf Initialization:} 
	the parameters of backbone $\theta$; the parameters of the following full connection layer $W$;\\
	\hspace*{0.02in} {\bf Optimization:} 
	\begin{algorithmic}[1]
		
		\For{number of training iterations} 
		\State Calculate CAAM and CAM;
		\State Update $W$ with $\triangledown _{W}L_{\text{ce}}$;
		\State Update $\theta$ with $\triangledown _{\theta }$CAM-loss
		\EndFor
		
		\State \Return optimal parameters $\theta^{*}$ and $W^{*}$
	\end{algorithmic}
	\label{alg:cam}
\end{algorithm}

\subsection{Choice of $\alpha$}
\label{chose}

\begin{figure}[h]
	\begin{center}
		\includegraphics[width=0.99\linewidth]{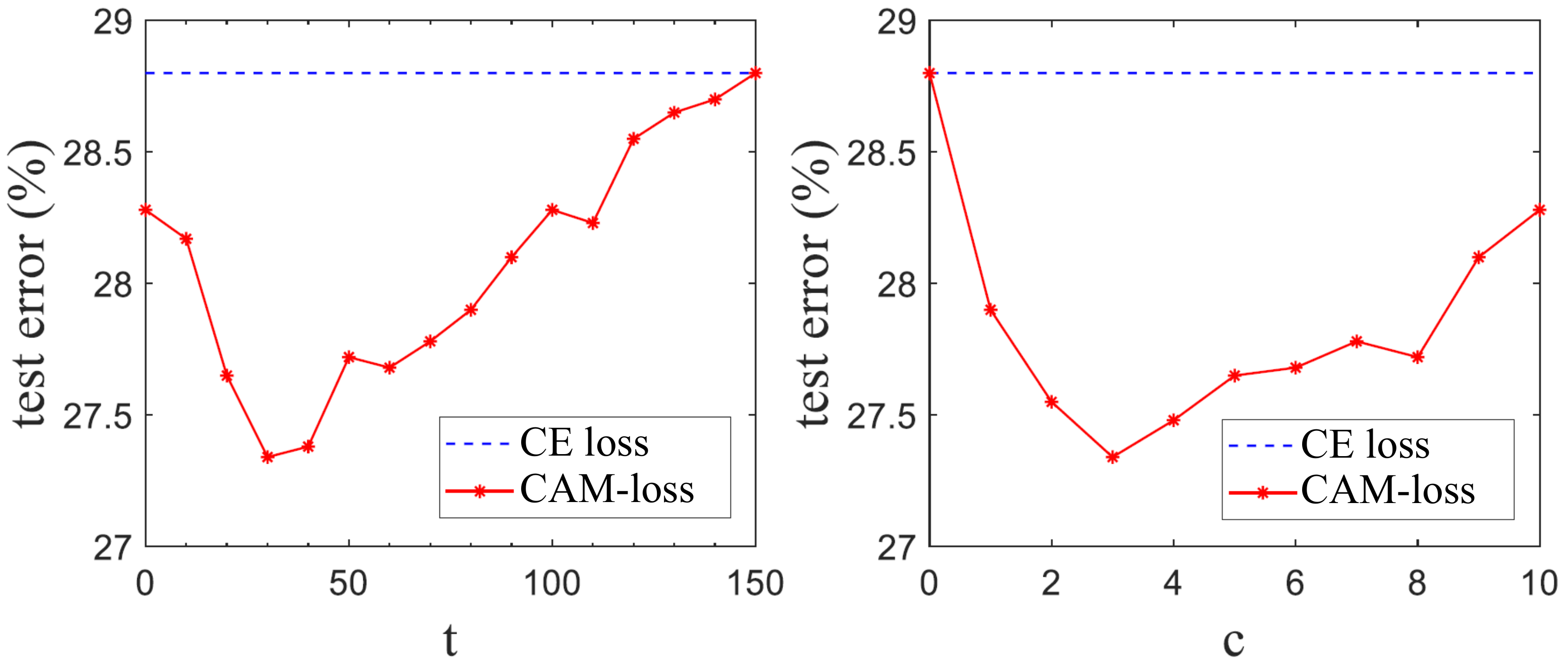}
	\end{center}
	\caption{Ablation of $t$ (left) and $c$ (right). We conduct an image classification experiment on CIFAR-100 with ResNet-56. The baseline adopts the cross entropy loss while our method adopts CAM-loss. The experimental setting can be seen in Sec. \ref{sec41}}
	\label{fig:alpha}
	\vspace{-2ex}
\end{figure}

How to choose $\alpha$ is an open question. Intuitively, CAMs obtained in the headmost epochs are too discrete to guide the CAAMs, while CAMs obtained in the latter epochs are more effective for guiding. So, we consider $\alpha$ as a simple step function formally described as follows
\begin{equation}
\alpha = \begin{cases}
0,  x<  t \\ 
c,  x \geqslant  t \end{cases},
\end{equation}
where $t$ is the jump point (or start epoch). It means that $L_{\text{cam}}$ will be added to $L_{\text{ce}}$ from the $t^{th}$ epoch. We simply set $c = 1$ to analyze the relationship between the value of $t$ and the error rate as shown in the left part of \autoref{fig:alpha}. It shows that the best $t$ is 30. Further analysis, we find that the train error and test error are approximately less than $50\%$ at the $30^{th}$ epoch. At the moment, CAMs already have obvious target category features. In this sense, adaptively setting the value of $t$ as the epoch when training accuracy exceeds $50\%$ is a simple but smart choice.

With the position of the start epoch $t$ fixed, we analyze the effect of the size of $c$ on the error rate as shown in the right part of \autoref{fig:alpha}. It shows that the improvement of error rate is maximal when $c = 3$. In fact, the choice of the optimal $\alpha$ is related to the dataset and the number of training epoch. $\alpha$ can also be any other function or a kind of probability distribution. It is difficult to traverse all possibilities, but we can always beat the baseline with a simple selection strategy as shown in \autoref{fig:alpha}.

\subsection{CAAM-CAM matching}

The classical knowledge distillation (KD\cite{hinton2015distilling}) lets a weak student mimic a strong teacher’s behavior by minimizing the Kullback-Leibler divergence of their soft targets. The formal description is shown below.

Given a vector of logits $z$ as the output of a deep model (or the input to the softmax), such that $z_{i}$ is the logit for the class $i$, and then the probability $p_{i}$ that the input image belongs to the class $i$ can be estimated by a softmax function, $p_{i}=\frac{e^{(z_{i})}}{\sum _{j}e^{(z_{j})}}$. A temperature factor $\tau$ is introduced to control the importance of each soft target as $p_{i}^{\tau}=\frac{e^{(z_{i}/\tau)}}{\sum _{j}e^{(z_{j}/\tau)}}$, where a higher temperature produces a softer probability distribution over classes. The distillation loss term of KD\cite{hinton2015distilling} is

\begin{equation}
L_{\text{kd}}=\frac{1}{n}\sum_{i=1}^{n}\tau^{2}(p_{ti}^{\tau}\text{log}p_{ti}^{\tau}-p_{ti}^{\tau}\text{log}p_{si}^{\tau}),
\end{equation}

where $p_{si}^{\tau}$ and $p_{ti}^{\tau}$ represent the soft targets $p_{i}^{\tau}$ of the student and teacher respectively. $n$ is the number of classes.

AT\cite{zagoruyko2016paying} proposes to match the attention maps between two different models. Using only the CAMs generated by the last convolutional layer, the distillation loss term of AT\cite{zagoruyko2016paying} can be simplified as follows
\begin{equation}
L_{\text{at}}=\left \| \text{CAM}_{si}^{'}-\text{CAM}_{ti}^{'} \right \|_{l_{1}},
\label{equ:at}
\end{equation}
where $\text{CAM}_{si}^{'}$ and $\text{CAM}_{ti}^{'}$ represent the normalized CAM of the student and teacher to the target class $i$ respectively. $l_{1}$ distance is used instead of $l_{2}$ to keep consistent.

Different from AT\cite{zagoruyko2016paying}, CAAM-CAM matching (CCM) adopts the normalized CAM of the target category generated by the teacher to constrain the normalized CAAM generated by the student. The CCM distillation loss term is
\begin{equation}
L_{\text{ccm}}=\left \| \text{CAAM}_{s}^{'}-\text{CAM}_{ti}^{'} \right \|_{l_{1}},
\label{equ:ccm}
\end{equation}
where $\text{CAAM}_{s}^{'}$ represents the normalized CAAM of the student. It's easy to find the difference between Eq. (\ref{equ:at}) and Eq. (\ref{equ:ccm}) is that  $\text{CAAM}_{s}^{'}$ replaces $\text{CAM}_{si}^{'}$. Further, Eq. (\ref{equ:ccm}) can be transformed as follows
\begin{footnotesize}
	\begin{equation}
	\begin{aligned}
	L_{\text{ccm}}&=\left \| \text{CAAM}_{s}^{'}-\text{CAM}_{ti}^{'} \right \|_{l_{1}} \\ 
	&\leqslant  \left \| \text{CAAM}_{s}^{'}-\text{CAM}_{si}^{'}\right \|_{l_{1}}+\left \|\text{CAM}_{si}^{'}-\text{CAM}_{ti}^{'} \right \|_{l_{1}} \\
	&=L_{\text{cam}}+L_{\text{at}}.
	\end{aligned}
	\label{equ:ccme}
	\end{equation}
\end{footnotesize}
\vspace{-1ex}

Eq. (\ref{equ:ccme}) states that $L_{\text{cam}}+L_{\text{at}}$ is the upper bound of $L_{\text{ccm}}$. In other words, CCM can be approximated as adding CAM-loss to the student network on the basis of AT\cite{zagoruyko2016paying}. Because CAM-loss can independently improve the performance of the student network, it is reasonable to infer that CCM can obtain better performance than AT\cite{zagoruyko2016paying}.

\begin{table*}[t]
	\begin{center}
		\begin{tabular}{lclc}
			\hline
			\multicolumn{2}{c}{CIFAR-100}                                                                                                                                                     & \multicolumn{2}{c}{ImageNet}                                                                                                                               \\ \hline
			Model                         & \begin{tabular}[c]{@{}c@{}}top 1\end{tabular}  & Model                         & \begin{tabular}[c]{@{}c@{}}top 1\end{tabular}  \\ \hline
			ResNet-56\cite{he2016deep}                     & 28.80                                                                                                                   & ResNet-50\cite{he2016deep}                     & 23.68                                                                                     \\
			ResNet-56\cite{he2016deep} + CAM-loss          & \textbf{27.34}                                                                                   & ResNet-50\cite{he2016deep} + CAM-loss          & \textbf{22.98}                                  \\ \hline
			Wide-ResNet-28-10\cite{zagoruyko2016wide}             & 18.37                                           & ResNet-101\cite{he2016deep}                    & 22.30       \\
			Wide-ResNet-28-10\cite{zagoruyko2016wide} + CAM-loss  & \textbf{17.49}             & ResNet-101\cite{he2016deep} + CAM-loss         & \textbf{21.73}                           \\ \hline
			ResNext-29, 8$\times$64d\cite{xie2017aggregated}             & 18.01                        & ResNext-50, 4$\times$32d\cite{xie2017aggregated}             & 22.42           \\
			ResNext-29, 8$\times$64d\cite{xie2017aggregated} + CAM-loss  & \textbf{17.24}               & ResNext-50, 4$\times$32d\cite{xie2017aggregated} + CAM-loss  & \textbf{21.91}    \\ \hline
			DenseNet-bc-190-40\cite{huang2017densely}            & 17.67            & ResNext-101, 8$\times$32d\cite{xie2017aggregated}            & 21.04           \\
			DenseNet-bc-190-40\cite{huang2017densely} + CAM-loss & \textbf{16.98}              & ResNext-101, 8$\times$32d\cite{xie2017aggregated} + CAM-loss & \textbf{20.45}           \\ \hline
		\end{tabular}
	\end{center}
	\caption{Applicability of CAM-loss to different network structures. Top 1 error rate (\%) is adopted, and results of CAM-loss are \textbf{bold-faced}.}
	\label{table:classification2}
\end{table*}
\begin{table*}[t]
	\begin{center}
		\begin{tabular}{lcclcc}
			\hline
			\multicolumn{3}{c}{CIFAR-100}                  & \multicolumn{3}{c}{ImageNet}                                                                                                                                                                   \\ \hline
			\begin{tabular}[c]{@{}l@{}}Baseline Model\\ PyramidNet-200 (alpha=240)\cite{han2017deep}\end{tabular} & \begin{tabular}[c]{@{}c@{}}top 1\end{tabular} &  \begin{tabular}[c]{@{}c@{}}top 5\end{tabular} & \begin{tabular}[c]{@{}l@{}}Baseline Model\\ ResNet-50\cite{he2016deep}\end{tabular} & \begin{tabular}[c]{@{}c@{}}top 1\end{tabular} & \begin{tabular}[c]{@{}c@{}}top 5\end{tabular} \\ \hline
			CE loss                                                                           & 16.45                                                        &  3.69                                                         & CE loss                                                          & 23.68                                                        & 7.05                                                         \\
			CAM-loss                                                                           & \textbf{15.79}                                               &  \textbf{3.28}                                                & CAM-loss                                                          & \textbf{22.98}                                               & \textbf{6.52}                                                \\ \hline
			Cutout\cite{devries2017improved}                                                                             & 16.53                                                        &  3.65                                                        & Cutout\cite{devries2017improved}                                                            & 22.93                                                        & 6.66                                                         \\ \hline
			Manifold Mixup\cite{verma2018manifold}                                                                     & 16.14                                                        &  4.07                                                         &Manifold Mixup\cite{verma2018manifold}                                                    & 22.50                                                        & 6.21                                                         \\ \hline
			StochDepth\cite{huang2016deep}                                                                         & 15.86                                                        &  3.33                                                         & 	StochDepth\cite{huang2016deep}                                                        & 22.46                                                        & 6.27                                                         \\ \hline
			DropBlock\cite{ghiasi2018dropblock}                                                                          & 15.73                                                        & 3.26                                                                                & DropBlock\cite{ghiasi2018dropblock}                                                        & 21.87                                                        & 5.98                                                         \\ \hline
			Mixup\cite{zhang2017mixup}                                                                              & 15.63                                                        & 3.99                                                                                & Mixup\cite{zhang2017mixup}                                                             & 22.58                                                        & 6.40                                                         \\ \hline
			Shakedrop\cite{yamada2018shakedrop}                                                                          & 15.08                                                        & 2.72                                                                                & \multicolumn{1}{c}{-}                                                                & -                                                        & -                                                         \\
			Shakedrop\cite{yamada2018shakedrop} + CAM-loss                                                               & \textbf{14.56}                                               & \textbf{2.56}                                                                       & \multicolumn{1}{c}{-}                                                                 & -                                                            & -                                                            \\ \hline
			Cutmix\cite{yun2019cutmix}                                                                             & 14.47                                                        & 2.97                                                                                & Cutmix\cite{yun2019cutmix}                                                            & 21.54                                                        & 5.92                                                         \\
			Cutmix\cite{yun2019cutmix} + CAM-loss                                                                  & \textbf{14.01}                                               & \textbf{2.93}                                                                       & Cutmix\cite{yun2019cutmix} + CAM-loss                                                 & \textbf{21.16}                                               & \textbf{5.79}                                                \\ \hline
			Cutmix + Shakedrop                                                                 & 13.81                                                        & 2.29                                                                                & \multicolumn{1}{c}{-}                                            & -                                                            & -                                                            \\
			Cutmix + Shakedrop + CAM-loss                                                      & \textbf{13.49}                                               & \textbf{2.18}                                                                       & \multicolumn{1}{c}{-}                                            & -                                                            & -                                                            \\ \hline
		\end{tabular}
	\end{center}
	\caption{Combination with mainstream regularization methods on CIFAR-100 and ImageNet. Top 1 and Top 5 error rate (\%) are adopted, and results of CAM-loss are \textbf{bold-faced}. Baseline results are obtained from\cite{yun2019cutmix}}
	\label{table:classification1}
	\vspace{-2ex}
\end{table*}

In practical application, like AT\cite{zagoruyko2016paying}, CCM also needs to combine the cross entropy loss term and soft target loss term to achieve the best performance. The loss function is
\begin{equation}
L =\beta L_{\text{ce}} + (1-\beta) L_{\text{kd}} + \gamma L_{\text{ccm}},
\label{equ:ccmloss}
\end{equation}
where $\beta$ and $\gamma$ represent the combine ratio. Because the teacher is a well-trained network that generates good CAMs, the optimization of $\beta$ and $\gamma$ can be done directly by linear search, without considering the influence of epoch like $\alpha$ in Sec.\ref{chose}.
\section{Experiments}
\label{others}

In this section, we first introduce the datasets in our experiments (Sec. \ref{sec40}). Then we evaluate the performance of CAM-loss in image classification tasks (Sec. \ref{sec41}), including application to various networks, combination with mainstream regularization methods, and comparison with different loss functions. We also verify the generalization ability of CAM-loss in transfer learning and few shot learning tasks (Sec. \ref{sec42}, \ref{sec43}). Finally, we apply CAAM-CAM matching method to knowledge distillation tasks (Sec. \ref{sec45}).

\subsection{Datasets}
\label{sec40}

\textbf{CIFAR-10 and CIFAR-100\cite{krizhevsky2009learning}.} The CIFAR-10 and CIFAR-100 comprise $32\times32$ pixel RGB images with 10 and 100 classes, containing 50,000 training and 10,000 test images. We follow the standard augmentation in\cite{howard2013some}. That is, the training images are padded 4 pixels, and then randomly cropped to $32\times32$ combined with random horizontal flipping. The original $32\times32$ images are used for testing.

\textbf{ImageNet-1K\cite{deng2009imagenet} and Mini-ImageNet\cite{vinyals2016matching}.} The ImageNet-1K contains 1.2 million training and 50,000 validation images of 1000 classes. The Mini-ImageNet consists of a subset of 100 classes from the ImageNet and contains 600 images for each class. We adopt the same augmentation strategy as\cite{yun2019cutmix} and apply a center cropping in testing. In the few shot learning task, following\cite{li2019revisiting}, we randomly split Mini-ImageNet\cite{vinyals2016matching} dataset into 64 base, 16 validation, and 20 novel classes.

\textbf{CUB-200-2011\cite{wah2011caltech} and Stanford Dogs\cite{khosla2011novel}.} The bird dataset contains 5,994 training and 5,794 testing images of 200 classes. The dog dataset contains 12,000 training and 8,580 testing images of 120 classes. For the data augmentation strategy, we rescale the input images to the resolution of $600\times600$, randomly crop a $448\times448$ region, and apply a center cropping in testing. In the few shot learning task, following \cite{li2019revisiting}, we randomly split the bird dataset into 120 base, 30 validation, and 50 novel classes.

\subsection{Image Classification}
\label{sec41}

We evaluate the image classification performances of CAM-loss on three benchmark datasets: CIFAR-10, CIFAR-100 and ImageNet-1K. On CIFAR datasets, we run 160 epochs with batch size 128, initial learning rate 0.1 and cosine learning rate decay. On ImageNet, we run 120 epochs with batch size 1024, initial learning rate 0.4 (batch size 512 and learning rate 0.2 for ResNext-101 due to computation limit) and cosine learning rate decay. Specially we set $c = 3$ and $t = 20$ (In fact, due to the robustness of CAM-loss to hyper parameters, we simply but confidently adopt the same setting in subsequent experiments).

\begin{figure}[h]
	\vspace{-3ex}
	\begin{center}
		\includegraphics[width=0.99\linewidth]{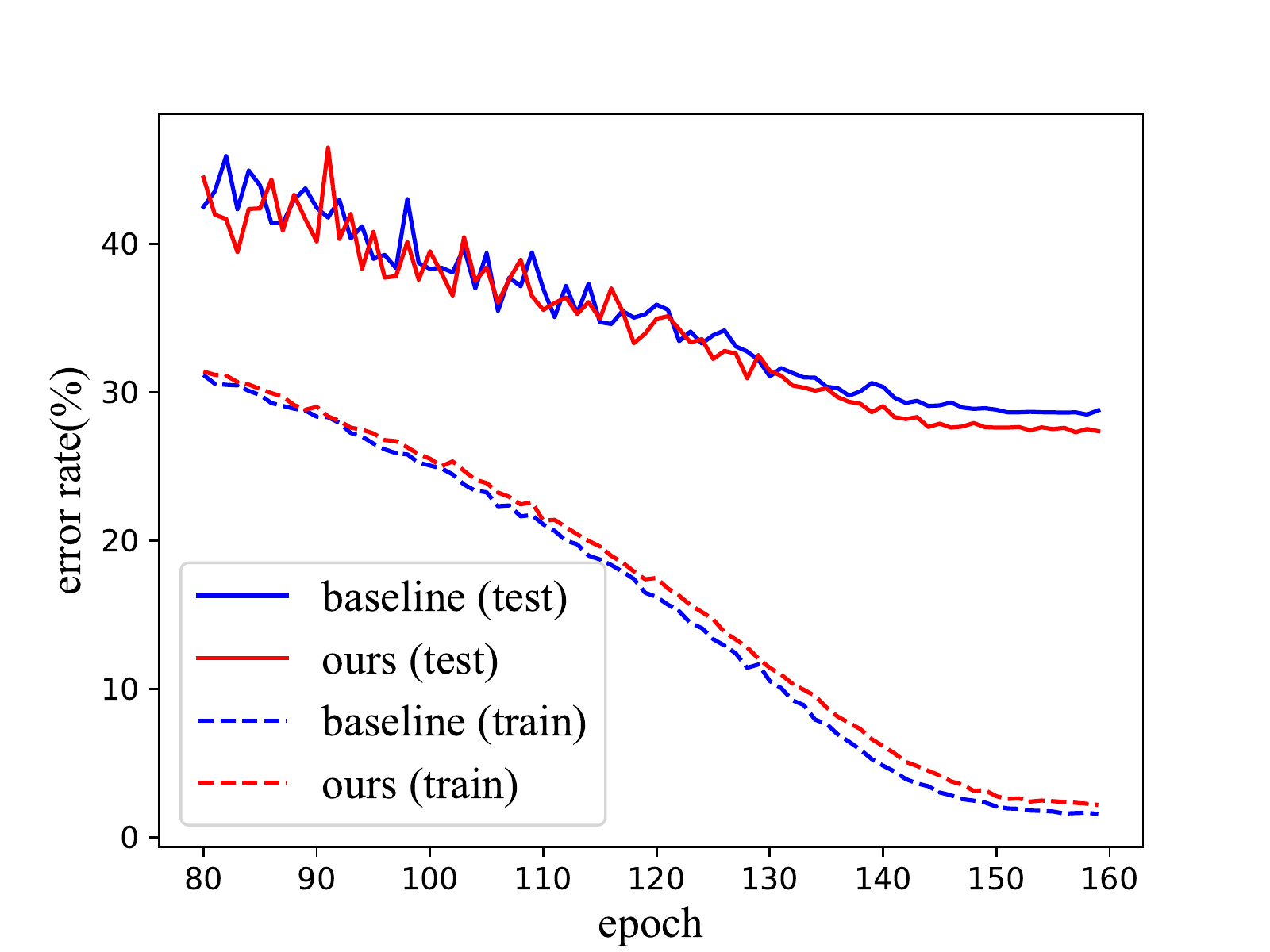}
	\end{center}
	\vspace{-10pt}
	\caption{Error rate vs. epoch with CAM-loss and cross entropy loss. We conduct an image classification experiment on CIFAR-100 with a ResNet-56 backbone. The baseline adopts the cross entropy loss while our method adopts CAM-loss.}
	\label{fig:overfit}
	\vspace{-2ex}
\end{figure}

\textbf{Apply to various network structures.} We perform ResNet \cite{he2016deep}, Wide-ResNet \cite{zagoruyko2016wide}, ResNext \cite{xie2017aggregated} and DenseNet \cite{huang2017densely} keeping all hyper parameters the same as original papers. \autoref{table:classification2} shows that CAM-loss can be widely used in a variety of network structures to improve the performances of baselines. Specifically, CAM-loss has brought 0.51-0.70\% improvement on ImageNet and 0.69-1.46\% improvement on CIFAR-100, which are significant under these large network structures. For further analysis, we focus on the relationship between the number of epoch and the error rate as shown in \autoref{fig:overfit}. It is observed that the model trained with CAM-loss achieves higher train error but lower test error, which proves that CAM-loss has a positive effect on avoiding overfitting. 

\textbf{Combine with regularization methods.} \autoref{table:classification1} shows the evaluation of different regularization methods on CIFAR-100 and ImageNet following the setup of\cite{yun2019cutmix}. We observe that CAM-loss can be widely combined with the mainstream regularization methods to improve their performances further. Specifically, CAM-loss reduces the top-1 error rate of ShakeDrop\cite{yamada2018shakedrop} by 0.52\% and CutMix\cite{yun2019cutmix} by 0.46\%. Surprisingly, it also reduces the top-1 error rate of the combination ShakeDrop\cite{yamada2018shakedrop} and CutMix\cite{yun2019cutmix} by 0.32\%. It means that CAM-loss can further boost the state-of-the-art regularization methods. There is no conflict between CAM-loss and most regularization methods, which is very welcome in practical applications.

\textbf{Compare with other loss functions.} In order to compare CAM-loss with popular loss functions, we conduct classification experiments with ResNet-110\cite{he2016deep} and Wide-ResNet-28-10\cite{zagoruyko2016wide} on CIFAR-10 and CIFAR-100 following the settings of \cite{wang2019implicit}. \autoref{table:classification3} shows that CAM-loss outperforms all of baseline loss functions. In fact, the routes of CAM-loss and other loss functions are parallel without conflict. They can be used together in an appropriate combine ratio setting.
In particular, we compare NegativeCAM \cite{sun2020fixing} with CAM-loss. ResNet-56\cite{he2016deep} and ResNet-110\cite{he2016deep} are adopted on CIFAR-100, and ResNet-50\cite{he2016deep} and ResNet-101\cite{he2016deep} are adopted on ImageNet. We inherit the loss module of Negative-CAM\cite{sun2020fixing} according to the official codes and reproduce the results by utilizing our baseline models.  \autoref{table:negativecam} shows that CAM-loss consistently outperforms NegativeCAM\cite{sun2020fixing}, especially on ImageNet, which means that CAM-loss is more efficient on complex datasets.
\begin{table}[h]
	\footnotesize
	\begin{center}
		\setlength{\tabcolsep}{0.6mm}{
		\begin{tabular}{lcccc}
			\hline
			\multicolumn{1}{c}{}                         & \multicolumn{2}{c}{ CIFAR-10}                                   & \multicolumn{2}{c}{ CIFAR-100}            \\
			\multicolumn{1}{l}{\multirow{-2}{*}{{ Method}}} & { ResNet-110}           & { WRN-28-10}             & { ResNet-110} & { WRN-28-10}    \\ \hline
				{ CE loss}                                  & { 6.76}        & { 3.82}          & { 27.68}            & {18.53}             \\
				{ center loss \cite{wen2016discriminative}}                                   & { 6.38}        & { 3.76}          & { 26.88} & { 18.50} \\
				{ L-softmax \cite{liu2016large}}                                     & { 6.46}        & { 3.69}          & { 27.03} & { 18.48} \\
				{ SM-softmax \cite{liang2017soft}}                                    & { 6.49}        & { 3.71}          & { 26.97} & { 18.40} \\ \hline
				{ CAM-loss (ours)}                               & { \textbf{6.29}} & { \textbf{3.49}} & { \textbf{26.56}}   & { \textbf{17.87}}    \\ \hline
		\end{tabular}}
	\end{center}
	\caption{Comparison with other loss functions. Top 1 error rate (\%) is adopted. Results of CAM-loss are \textbf{bold-faced}.}
	\label{table:classification3}
	\vspace{-1ex}
\end{table}

\begin{table}[h]
	\footnotesize
	\begin{center}
		\setlength{\tabcolsep}{0.8mm}{
		\begin{tabular}{lcccc}
			\hline
			\multicolumn{1}{l}{\multirow{2}{*}{Methods}} & \multicolumn{2}{c}{CIFAR-100} & \multicolumn{2}{c}{ImageNet} \\
			\multicolumn{1}{c}{}                         & ResNet-56     & ResNet-110    & ResNet-50    & ResNet-101    \\ \hline
			CE loss                                           & 28.80         & 27.68         & 23.68        & 22.30         \\
			NegativeCAM\cite{sun2020fixing}                                  & 27.37         & 26.76         & 23.32        & 22.02         \\ \hline
			CAM-loss                                     & \textbf{27.34}         & \textbf{26.56}         & \textbf{22.98}        & \textbf{21.73}         \\ \hline
		\end{tabular}}
	\end{center}

\caption{Comparison with NegativeCAM \cite{sun2020fixing} on CIFAR-100 and ImageNet. Top 1 error rate (\%) is adopted. Results of CAM-loss are \textbf{bold-faced}.}
\label{table:negativecam}
\end{table}

\subsection{Transfer learning}

\label{sec42}
We evaluate the generalization ability of CAM-loss under the setting of transfer learning. A ResNet-50\cite{he2016deep} model pre-trained with CAM-loss on ImageNet-1K\cite{deng2009imagenet}, and then finetune it on CUB\cite{wah2011caltech} and Stanford Dogs\cite{khosla2011novel}. Following\cite{dubey2018pairwise,sun2018multi}, we finetune the pre-trained model on the training set for $90$ epochs with batch size $8$ (CUB) and $16$ (Stanford Dogs). SGD optimizer is adopted with an initial learning rate $0.001$ and a cosine learning rate decay. We set the weight decay as $5\times10^{-4}$ and momentum as $0.9$. \autoref{table:transfer} shows that CAM-loss improves the baseline by $1.1\%$ and $0.7\%$ on CUB\cite{wah2011caltech} and Stanford Dogs\cite{khosla2011novel}. It confirms that CAM-loss has stronger classification capability on novel dataset compared with the cross entropy loss.
\begin{table}[h]
	\begin{center}
		\setlength{\tabcolsep}{3mm}{
		\begin{tabular}{lcc}
			\hline
			Dataset       &CE loss &CAM-loss\\ \hline
			CUB \cite{wah2011caltech}  & 85.6                                                              & \textbf{86.7}                                                                         \\ \hline
			Stanford Dogs\cite{khosla2011novel} & 83.9                                                              & \textbf{84.6}                                                                         \\ \hline
		\end{tabular}}
	\end{center}
	\caption{Accuracies of fine-grained classification tasks under the setting of transfer learning. Top 1 accuracy (\%) is adopted. Results of CAM-loss are \textbf{bold-faced}.}
	\label{table:transfer}
\end{table}

\begin{table*}[h]
	\begin{center}
		\begin{tabular}{lccccc}
			\hline
			\multirow{2}{*}{Method} & \multicolumn{2}{c}{Mini-ImageNet} & \multicolumn{2}{c}{CUB} & Mini$\rightarrow$CUB \\ \cline{2-6} 
			& 1-shot           & 5-shot          & 1-shot          & 5-shot         & 5-shot                 \\ \hline
			Baseline++\cite{chen2019closer}               & 52.18       & 75.86      & 67.08      & 84.19     & 65.88             \\ \hline
			Baseline++\cite{chen2019closer} + CAM-loss    &\textbf{54.80}       & \textbf{77.54}      & \textbf{74.12}      & \textbf{88.93}     & \textbf{68.63}             \\ \hline
		\end{tabular}
	\end{center}
	\caption{Accuracies of few shot classification tasks. Mini$\rightarrow$CUB represents the cross-domain. Results of CAM-loss are \textbf{bold-faced}.}
	\label{table:fewshot}
\end{table*}

\begin{table*}[h]
	\begin{center}
		\begin{tabular}{cllcccc}
			\hline
			Setup      & Teacher                               & Student   & \begin{tabular}[c]{@{}l@{}}baseline\end{tabular} & KD\cite{hinton2015distilling} & AT\cite{zagoruyko2016paying}    & CCM \\ \hline
			(a)      &  WRN-28-4\cite{zagoruyko2016wide}       & WRN-16-4\cite{zagoruyko2016wide}  & 23.14                                                          & 21.93    & 21.77 & \textbf{21.46}         \\ \hline
			(b)     &  WRN-28-4\cite{zagoruyko2016wide}       & WRN-28-2\cite{zagoruyko2016wide}  & 25.40                                                           & 23.12    & 22.82 & \textbf{22.50}         \\ \hline
			(c)      &  WRN-28-4\cite{zagoruyko2016wide}      & WRN-16-2\cite{zagoruyko2016wide}  & 27.94                                                          & 26.05    & 25.85 & \textbf{25.45}         \\ \hline
			(d)      &  WRN-28-4\cite{zagoruyko2016wide}        & Resnet-56\cite{he2016deep} & 28.80                                                           & 27.11    & 26.98 & \textbf{26.48}         \\ \hline
			(e)      &  PyramidNet-200\cite{han2017deep} & WRN-28-4\cite{zagoruyko2016wide}   & 20.97                                                          & 20.08    & 20.22 & \textbf{19.93}         \\ \hline
		\end{tabular}
	\end{center}
	\caption{Performance of various knowledge distillation setups on CIFAR-100. 'WRN' denotes Wide-ResNet for short. Baseline denotes the top 1 error rate (\%) of the student network. Results of CCM method are \textbf{bold-faced}.}
	\label{table:distil}
	
	\vspace{-2ex}
\end{table*}

\subsection{Few shot learning}
\label{sec43}

Due to limited data on novel classes, few shot learning relies heavily on the generalization ability of models trained on the base classes. The mainstream few shot image classification methods are evaluated and compared in\cite{chen2019closer}, in which Baseline++ is verified  with the competitive classification performance and generalization capability. For simplicity, we add CAM-loss to the Baseline++ method to evaluate the performance improvements under three scenarios: (1) general object recognition, (2) fine-grained image classification, (3) cross-domain adaptation (using Mini-ImageNet\cite{vinyals2016matching} as base classes and the 50 validation and 50 novel classes from CUB\cite{wah2011caltech}). We adopt the same setting as \cite{chen2019closer} except for the hyper parameters of CAM-loss.

\autoref{table:fewshot} shows that under the standard 5-way 1-shot and 5-shot protocols, CAM-loss averagely boosts Baseline++ by $7.04\%$ and $4.75\%$ on CUB\cite{wah2011caltech}, $2.78\%$ and $1.68\%$ on Mini-ImageNet\cite{vinyals2016matching}. The cross-domain result of Baseline++ is also significantly improved by $2.75\%$. As far as we know, our results of few shot classification on CUB\cite{wah2011caltech} has been very competitive with the state-of-the-art in inductive inference setting. 

\begin{figure}[h]
	\setlength{\belowcaptionskip}{-1.cm}
	\begin{center}
		\includegraphics[width=0.99\linewidth]{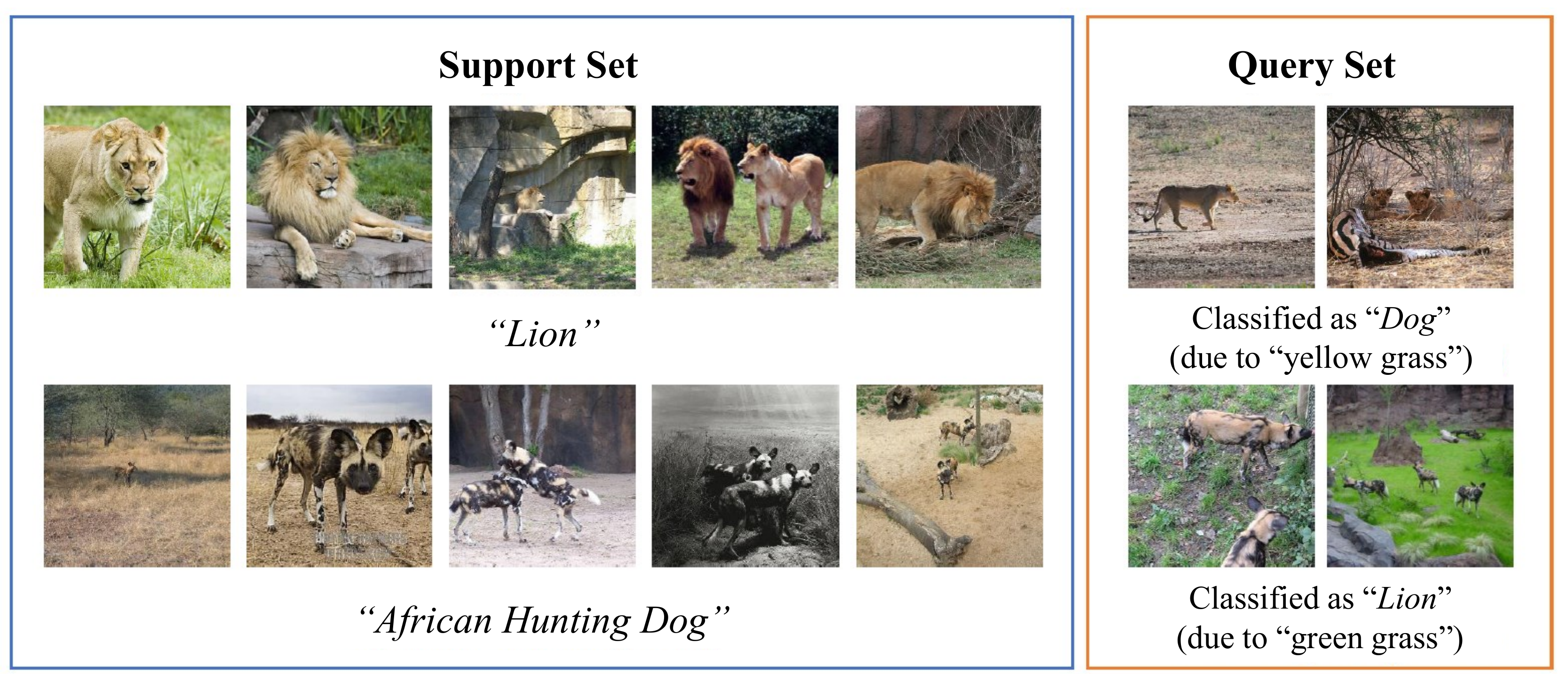}
	\end{center}
	\caption{Negative effects of background in few shot image classification.\cite{yue2020interventional}}
	\label{fig:fsl}
\end{figure}

Why does CAM-loss perform so well in few shot image classification tasks? \cite{hou2019cross, yue2020interventional} has repeatedly confirmed that the features of background bring great troubles to few shot image classification. A 5-shot learning example is shown in \autoref{fig:fsl}, where the features of yellow grass and green grass are misleading. In query set, the lion samples with yellow grass are misclassified as dog while the dog samples with green grass are misclassified as lion. CAM-loss is an effective method to suppress the expression of background features. That is very helpful to reduce the negative effect of background in few shot image classification tasks, especially in the fine-grained few shot image classification.

\subsection{Knowledge distillation}
\label{sec45}

We conduct knowledge distillation experiments on CIFAR-100\cite{krizhevsky2009learning}, and choose KD\cite{hinton2015distilling} and AT\cite{zagoruyko2016paying} as the baseline methods. For KD\cite{hinton2015distilling}, we set the hyper parameter temperature as $4$ and combine ratio of loss terms as $0.5$. For AT\cite{zagoruyko2016paying}, we choose the best strategy that is to set $(\beta)l_{\text{ce}}+(1-\beta)l_{\text{kd}}+\gamma l_{\text{at}}$ as loss function where $l_{\text{at}}$ adopts $l_{2}$ distance, $\beta$ is set as 0.5 and $\gamma$ is set as 10. For CCM, we set $\beta$ as 0.5, $\gamma$ as 1 following Eq. (\ref{equ:ccm}) and (\ref{equ:ccmloss}). For generality of the experiments, we adpoted various teacher/student pairs with the same depth (WRN-28-4/WRN-28-2), different depth (WRN-28-4/WRN-16-2, WRN-28-4/WRN-16-4), different type (WRN-28-4/ResNet-56, PyramidNet-200/WRN-28-4). \autoref{table:distil} shows that CCM consistently outperforms the two baseline methods. Further thinking, CAM-loss can also be seen as a self-distillation strategy, that is, the supervision information comes from the network itself rather than the teacher network.

\section{Conclusion}

In this paper, we have proposed a novel loss function CAM-loss to boost the performance of CNN classification models. Essentially, it constrains the feature maps with the spatial information from CAMs. A model trained with CAM-loss is inclined to express the features of target category and suppress those of non-target categories, which is effective to enforce intra-class compactness and inter-class separability. As an independent loss function, it can be easily combined with mainstream regularization methods to improve their performance in image classification tasks. Strong generalization capability makes it outstanding in transfer learning and few shot learning tasks. Based on CAM-loss, we also propose a novel CCM knowledge distillation method, which matches different knowledge between teacher and student. In future, we will study the applications of CAM-loss to more generic visual tasks.

\section*{Acknowledgments}

This work is supported in part by the National Science and Technology Major Project of the Ministry of Science and Technology of China under Grants 2018AAA0100701, the National Natural Science Foundation of China under Grants 61906106 and 62022048, and Beijing Academy of Artificial Intelligence.
{\small
\bibliographystyle{ieee_fullname}
\bibliography{egbib}

\begin{thebibliography}{10}\itemsep=-1pt

\bibitem{ahn2019weakly}
Jiwoon Ahn, Sunghyun Cho, and Suha Kwak.
\newblock Weakly supervised learning of instance segmentation with inter-pixel
  relations.
\newblock In {\em Proceedings of the IEEE Conference on Computer Vision and
  Pattern Recognition}, pages 2209--2218, 2019.

\bibitem{baek2020psynet}
Kyungjune Baek, Minhyun Lee, and Hyunjung Shim.
\newblock Psynet: Self-supervised approach to object localization using point
  symmetric transformation.
\newblock In {\em Proceedings of the AAAI Conference on Artificial
  Intelligence}, volume~34, pages 10451--10459, 2020.

\bibitem{chattopadhay2018grad}
Aditya Chattopadhay, Anirban Sarkar, Prantik Howlader, and Vineeth~N
  Balasubramanian.
\newblock Grad-cam++: Generalized gradient-based visual explanations for deep
  convolutional networks.
\newblock In {\em 2018 IEEE Winter Conference on Applications of Computer
  Vision}, pages 839--847, 2018.

\bibitem{chen2019closer}
Weiyu Chen, Yencheng Liu, Zsolt KiraAuthors, Yu Chiang, and Huang Jiabin.
\newblock A closer look at few-shot classification.
\newblock In {\em Proceedings of the IEEE International Conference on Learning
  Representations Worshops}, 2019.

\bibitem{deng2009imagenet}
Jia Deng, Wei Dong, Richard Socher, Li-Jia Li, Kai Li, and Li Fei-Fei.
\newblock Imagenet: A large-scale hierarchical image database.
\newblock In {\em 2009 IEEE Conference on Computer Vision and Pattern
  Recognition}, pages 248--255, 2009.

\bibitem{devries2017improved}
Terrance DeVries and Graham~W Taylor.
\newblock Improved regularization of convolutional neural networks with cutout.
\newblock {\em arXiv preprint arXiv:1708.04552}, 2017.

\bibitem{dubey2018pairwise}
Abhimanyu Dubey, Otkrist Gupta, Pei Guo, Ramesh Raskar, Ryan Farrell, and
  Nikhil Naik.
\newblock Pairwise confusion for fine-grained visual classification.
\newblock In {\em Proceedings of the European Conference on Computer Vision},
  pages 70--86, 2018.

\bibitem{fukui2019attention}
Hiroshi Fukui, Tsubasa Hirakawa, Takayoshi Yamashita, and Hironobu Fujiyoshi.
\newblock Attention branch network: Learning of attention mechanism for visual
  explanation.
\newblock In {\em Proceedings of the IEEE Conference on Computer Vision and
  Pattern Recognition}, pages 10705--10714, 2019.

\bibitem{ghiasi2018dropblock}
Golnaz Ghiasi, Tsung-Yi Lin, and Quoc~V Le.
\newblock Dropblock: A regularization method for convolutional networks.
\newblock In {\em Advances in Neural Information Processing Systems}, pages
  10727--10737, 2018.

\bibitem{goodfellow2016deep}
Ian Goodfellow, Yoshua Bengio, and Aaron Courville.
\newblock {\em Deep learning}.
\newblock MIT press, 2016.

\bibitem{guo2019visual}
Hao Guo, Kang Zheng, Xiaochuan Fan, Hongkai Yu, and Song Wang.
\newblock Visual attention consistency under image transforms for multi-label
  image classification.
\newblock In {\em Proceedings of the IEEE Conference on Computer Vision and
  Pattern Recognition}, pages 729--739, 2019.

\bibitem{hadsell2006dimensionality}
Raia Hadsell, Sumit Chopra, and Yann LeCun.
\newblock Dimensionality reduction by learning an invariant mapping.
\newblock In {\em 2006 IEEE Computer Society Conference on Computer Vision and
  Pattern Recognition}, volume~2, pages 1735--1742, 2006.

\bibitem{han2017deep}
Dongyoon Han, Jiwhan Kim, and Junmo Kim.
\newblock Deep pyramidal residual networks.
\newblock In {\em Proceedings of the IEEE Conference on Computer Vision and
  Pattern Recognition}, pages 5927--5935, 2017.

\bibitem{he2016deep}
Kaiming He, Xiangyu Zhang, Shaoqing Ren, and Jian Sun.
\newblock Deep residual learning for image recognition.
\newblock In {\em Proceedings of the IEEE Conference on Computer Vision and
  Pattern Recognition}, pages 770--778, 2016.

\bibitem{hinton2015distilling}
Geoffrey Hinton, Oriol Vinyals, and Jeff Dean.
\newblock Distilling the knowledge in a neural network.
\newblock {\em arXiv preprint arXiv:1503.02531}, 2015.

\bibitem{hou2019cross}
Ruibing Hou, Hong Chang, MA Bingpeng, Shiguang Shan, and Xilin Chen.
\newblock Cross attention network for few-shot classification.
\newblock In {\em Advances in Neural Information Processing Systems}, pages
  4003--4014, 2019.

\bibitem{howard2013some}
Andrew~G Howard.
\newblock Some improvements on deep convolutional neural network based image
  classification.
\newblock {\em arXiv preprint arXiv:1312.5402}, 2013.

\bibitem{huang2017densely}
Gao Huang, Zhuang Liu, Laurens Van Der~Maaten, and Kilian~Q Weinberger.
\newblock Densely connected convolutional networks.
\newblock In {\em Proceedings of the IEEE Conference on Computer Vision and
  Pattern Recognition}, pages 4700--4708, 2017.

\bibitem{huang2016deep}
Gao Huang, Yu Sun, Zhuang Liu, Daniel Sedra, and Kilian~Q Weinberger.
\newblock Deep networks with stochastic depth.
\newblock In {\em European Conference on Computer Vision}, pages 646--661.
  Springer, 2016.

\bibitem{khosla2011novel}
Aditya Khosla, Nityananda Jayadevaprakash, Bangpeng Yao, and Fei-Fei Li.
\newblock Novel dataset for fine-grained image categorization: Stanford dogs.
\newblock In {\em Proc. CVPR Workshop on Fine-Grained Visual Categorization},
  volume~2, 2011.

\bibitem{krizhevsky2009learning}
Alex Krizhevsky, Geoffrey Hinton, et~al.
\newblock Learning multiple layers of features from tiny images.
\newblock 2009.

\bibitem{krizhevsky2012imagenet}
Alex Krizhevsky, Ilya Sutskever, and Geoffrey~E Hinton.
\newblock Imagenet classification with deep convolutional neural networks.
\newblock In {\em Advances in Neural Information Processing Systems}, pages
  1097--1105, 2012.

\bibitem{li2018tell}
Kunpeng Li, Ziyan Wu, Kuan-Chuan Peng, Jan Ernst, and Yun Fu.
\newblock Tell me where to look: Guided attention inference network.
\newblock In {\em Proceedings of the IEEE Conference on Computer Vision and
  Pattern Recognition}, pages 9215--9223, 2018.

\bibitem{li2019revisiting}
Wenbin Li, Lei Wang, Jinglin Xu, Jing Huo, Yang Gao, and Jiebo Luo.
\newblock Revisiting local descriptor based image-to-class measure for few-shot
  learning.
\newblock In {\em Proceedings of the IEEE Conference on Computer Vision and
  Pattern Recognition}, pages 7260--7268, 2019.

\bibitem{liang2017soft}
Xuezhi Liang, Xiaobo Wang, Zhen Lei, Shengcai Liao, and Stan~Z Li.
\newblock Soft-margin softmax for deep classification.
\newblock In {\em International Conference on Neural Information Processing},
  pages 413--421. Springer, 2017.

\bibitem{liu2016large}
Weiyang Liu, Yandong Wen, Zhiding Yu, and Meng Yang.
\newblock Large-margin softmax loss for convolutional neural networks.
\newblock In {\em International Conference on Machine Learning}, volume~2,
  page~7, 2016.

\bibitem{peng2019correlation}
Baoyun Peng, Xiao Jin, Jiaheng Liu, Dongsheng Li, Yichao Wu, Yu Liu, Shunfeng
  Zhou, and Zhaoning Zhang.
\newblock Correlation congruence for knowledge distillation.
\newblock In {\em Proceedings of the IEEE International Conference on Computer
  Vision}, pages 5007--5016, 2019.

\bibitem{romero2014fitnets}
Adriana Romero, Nicolas Ballas, Samira~Ebrahimi Kahou, Antoine Chassang, Carlo
  Gatta, and Yoshua Bengio.
\newblock Fitnets: Hints for thin deep nets.
\newblock {\em arXiv preprint arXiv:1412.6550}, 2014.

\bibitem{russakovsky2015imagenet}
Olga Russakovsky, Jia Deng, Hao Su, Jonathan Krause, Sanjeev Satheesh, Sean Ma,
  Zhiheng Huang, Andrej Karpathy, Aditya Khosla, Michael Bernstein, et~al.
\newblock Imagenet large scale visual recognition challenge.
\newblock {\em International Journal of Computer Vision}, 115(3):211--252,
  2015.

\bibitem{schroff2015facenet}
Florian Schroff, Dmitry Kalenichenko, and James Philbin.
\newblock Facenet: A unified embedding for face recognition and clustering.
\newblock In {\em Proceedings of the IEEE Conference on Computer Vision and
  Pattern Recognition}, pages 815--823, 2015.

\bibitem{selvaraju2016grad}
Ramprasaath~R Selvaraju, Abhishek Das, Ramakrishna Vedantam, Michael Cogswell,
  Devi Parikh, and Dhruv Batra.
\newblock Grad-cam: Why did you say that?
\newblock {\em arXiv preprint arXiv:1611.07450}, 2016.

\bibitem{simonyan2014very}
Karen Simonyan and Andrew Zisserman.
\newblock Very deep convolutional networks for large-scale image recognition.
\newblock {\em arXiv preprint arXiv:1409.1556}, 2014.

\bibitem{srivastava2014dropout}
Nitish Srivastava, Geoffrey Hinton, Alex Krizhevsky, Ilya Sutskever, and Ruslan
  Salakhutdinov.
\newblock Dropout: a simple way to prevent neural networks from overfitting.
\newblock {\em The Journal of Machine Learning Research}, 15(1):1929--1958,
  2014.

\bibitem{sun2020fixing}
Guolei Sun, Salman Khan, Wen Li, Hisham Cholakkal, and Fahad Shahbaz.
\newblock Fixing localization errors to improve image classification.
\newblock 2020.

\bibitem{sun2018multi}
Ming Sun, Yuchen Yuan, Feng Zhou, and Errui Ding.
\newblock Multi-attention multi-class constraint for fine-grained image
  recognition.
\newblock In {\em Proceedings of the European Conference on Computer Vision},
  pages 805--821, 2018.

\bibitem{verma2018manifold}
Vikas Verma, Alex Lamb, Christopher Beckham, Amir Najafi, Ioannis Mitliagkas,
  Aaron Courville, David Lopez-Paz, and Yoshua Bengio.
\newblock Manifold mixup: Better representations by interpolating hidden
  states.
\newblock {\em arXiv preprint arXiv:1806.05236}, 2018.

\bibitem{vinyals2016matching}
Oriol Vinyals, Charles Blundell, Timothy Lillicrap, Daan Wierstra, et~al.
\newblock Matching networks for one shot learning.
\newblock In {\em Advances in Neural Information Processing Systems}, pages
  3630--3638, 2016.

\bibitem{wah2011caltech}
Catherine Wah, Steve Branson, Peter Welinder, Pietro Perona, and Serge
  Belongie.
\newblock The caltech-ucsd birds-200-2011 dataset.
\newblock 2011.

\bibitem{wan2018min}
Fang Wan, Pengxu Wei, Jianbin Jiao, Zhenjun Han, and Qixiang Ye.
\newblock Min-entropy latent model for weakly supervised object detection.
\newblock In {\em Proceedings of the IEEE Conference on Computer Vision and
  Pattern Recognition}, pages 1297--1306, 2018.

\bibitem{wang2018additive}
Feng Wang, Jian Cheng, Weiyang Liu, and Haijun Liu.
\newblock Additive margin softmax for face verification.
\newblock {\em IEEE Signal Processing Letters}, 25(7):926--930, 2018.

\bibitem{wang2021regularizing}
Yulin Wang, Gao Huang, Shiji Song, Xuran Pan, Yitong Xia, and Cheng Wu.
\newblock Regularizing deep networks with semantic data augmentation.
\newblock {\em IEEE Transactions on Pattern Analysis and Machine Intelligence},
  2021.

\bibitem{wang2019implicit}
Yulin Wang, Xuran Pan, Shiji Song, Hong Zhang, Gao Huang, and Cheng Wu.
\newblock Implicit semantic data augmentation for deep networks.
\newblock In {\em Advances in Neural Information Processing Systems}, pages
  12614--12623, 2019.

\bibitem{wei2018ts2c}
Yunchao Wei, Zhiqiang Shen, Bowen Cheng, Honghui Shi, Jinjun Xiong, Jiashi
  Feng, and Thomas Huang.
\newblock Ts2c: Tight box mining with surrounding segmentation context for
  weakly supervised object detection.
\newblock In {\em Proceedings of the European Conference on Computer Vision},
  pages 434--450, 2018.

\bibitem{wei2018revisiting}
Yunchao Wei, Huaxin Xiao, Honghui Shi, Zequn Jie, Jiashi Feng, and Thomas~S
  Huang.
\newblock Revisiting dilated convolution: A simple approach for weakly-and
  semi-supervised semantic segmentation.
\newblock In {\em Proceedings of the IEEE Conference on Computer Vision and
  Pattern Recognition}, pages 7268--7277, 2018.

\bibitem{wen2016discriminative}
Yandong Wen, Kaipeng Zhang, Zhifeng Li, and Yu Qiao.
\newblock A discriminative feature learning approach for deep face recognition.
\newblock In {\em European Conference on Computer Vision}, pages 499--515.
  Springer, 2016.

\bibitem{xie2017aggregated}
Saining Xie, Ross Girshick, Piotr Doll{\'a}r, Zhuowen Tu, and Kaiming He.
\newblock Aggregated residual transformations for deep neural networks.
\newblock In {\em Proceedings of the IEEE Conference on Computer Vision and
  Pattern Recognition}, pages 1492--1500, 2017.

\bibitem{yamada2018shakedrop}
Yoshihiro Yamada, Masakazu Iwamura, Takuya Akiba, and Koichi Kise.
\newblock Shakedrop regularization for deep residual learning.
\newblock {\em arXiv preprint arXiv:1802.02375}, 2018.

\bibitem{yang2020combinational}
Seunghan Yang, Yoonhyung Kim, Youngeun Kim, and Changick Kim.
\newblock Combinational class activation maps for weakly supervised object
  localization.
\newblock In {\em The IEEE Winter Conference on Applications of Computer
  Vision}, pages 2941--2949, 2020.

\bibitem{yim2017gift}
Junho Yim, Donggyu Joo, Jihoon Bae, and Junmo Kim.
\newblock A gift from knowledge distillation: Fast optimization, network
  minimization and transfer learning.
\newblock In {\em Proceedings of the IEEE Conference on Computer Vision and
  Pattern Recognition}, pages 4133--4141, 2017.

\bibitem{yue2020interventional}
Zhongqi Yue, Hanwang Zhang, Qianru Sun, and Xian-Sheng Hua.
\newblock Interventional few-shot learning.
\newblock {\em arXiv preprint arXiv:2009.13000}, 2020.

\bibitem{yun2019cutmix}
Sangdoo Yun, Dongyoon Han, Seong~Joon Oh, Sanghyuk Chun, Junsuk Choe, and
  Youngjoon Yoo.
\newblock Cutmix: Regularization strategy to train strong classifiers with
  localizable features.
\newblock In {\em Proceedings of the IEEE International Conference on Computer
  Vision}, pages 6023--6032, 2019.

\bibitem{zagoruyko2016paying}
Sergey Zagoruyko and Nikos Komodakis.
\newblock Paying more attention to attention: Improving the performance of
  convolutional neural networks via attention transfer.
\newblock {\em arXiv preprint arXiv:1612.03928}, 2016.

\bibitem{zagoruyko2016wide}
Sergey Zagoruyko and Nikos Komodakis.
\newblock Wide residual networks.
\newblock {\em arXiv preprint arXiv:1605.07146}, 2016.

\bibitem{zhang2017mixup}
Hongyi Zhang, Moustapha Cisse, Yann~N Dauphin, and David Lopez-Paz.
\newblock mixup: Beyond empirical risk minimization.
\newblock {\em arXiv preprint arXiv:1710.09412}, 2017.

\bibitem{zhang2018zigzag}
Xiaopeng Zhang, Jiashi Feng, Hongkai Xiong, and Qi Tian.
\newblock Zigzag learning for weakly supervised object detection.
\newblock In {\em Proceedings of the IEEE Conference on Computer Vision and
  Pattern Recognition}, pages 4262--4270, 2018.

\bibitem{zhou2016learning}
Bolei Zhou, Aditya Khosla, Agata Lapedriza, Aude Oliva, and Antonio Torralba.
\newblock Learning deep features for discriminative localization.
\newblock In {\em Proceedings of the IEEE Conference on Computer Vision and
  Pattern Recognition}, pages 2921--2929, 2016.

\end{thebibliography}
}

\end{document}